\def\year{2019}\relax
\documentclass[letterpaper]{article} 
\usepackage{aaai19}  
\usepackage{times}  
\usepackage{helvet}  
\usepackage{courier}  
\usepackage{url}  
\usepackage{graphicx}  

\usepackage{amsmath}
\usepackage{dcolumn}
\usepackage{kotex}
\usepackage{color}
\usepackage{hhline}
\usepackage{verbatim}
\usepackage{multirow}
\usepackage{amssymb}
\usepackage{rotating}
\usepackage{subfigure}
\usepackage{dblfloatfix}
\usepackage{enumitem}

\newcolumntype{L}[1]{>{\raggedright\let\newline\\\arraybackslash\hspace{0pt}}m{#1}}
\newcolumntype{C}[1]{>{\centering\let\newline\\\arraybackslash\hspace{0pt}}m{#1}}
\newcolumntype{R}[1]{>{\raggedleft\let\newline\\\arraybackslash\hspace{0pt}}m{#1}}

\newcommand\Tstrut{\rule{0pt}{2.5ex}}         

\begin{document}
%
\title{Detecting Incongruity Between News Headline and Body Text \\
via a Deep Hierarchical Encoder}

\author{Seunghyun Yoon$^{1,2}$,~Kunwoo Park$^{3,4}$,~Joongbo Shin$^{1}$,~Hongjun Lim$^{4}$\\
\bf \Large Seungpil Won$^{1}$,~Meeyoung Cha$^{5,4}$~\and~Kyomin Jung$^{1,2}$\\
$^{1}$Department of Electrical and Computer Engineering, Seoul National University, Seoul,  Korea\\
$^{2}$Automation and Systems Research Institute, Seoul National University, Seoul, Korea \\
$^{3}$Qatar Computing Research Institute, Doha, Qatar~~~~~
$^{4}$School of Computing, KAIST, Daejeon,  Korea\\
$^{5}$Pioneer Research Center for Mathematical and Computational Sciences, Institute for Basic Science (IBS), Daejeon, Korea\\
}


\maketitle

\newcommand{\kw}[1]{\textcolor{red}{\small [ToDo: #1 (left by Kunwoo)]}}
\newcommand{\kwq}[1]{\textcolor{red}{\small [Q: #1 (left by Kunwoo)]}}
\newcommand{\kwc}[1]{\textcolor{red}{\small [Comment: #1 (left by Kunwoo)]}}
\newcommand{\kwm}[1]{\textcolor{black}{#1}}
\newcommand{\sh}[1]{\textcolor{blue}{#1}}

\begin{abstract}
Some news headlines mislead readers with overrated or false information, and identifying them in advance will better assist readers in choosing proper news stories to consume. This research introduces million-scale pairs of news headline and body text dataset with incongruity label, which can uniquely be utilized for detecting news stories with misleading headlines. On this dataset, we develop two neural networks with hierarchical architectures that model a complex textual representation of news articles and measure the incongruity between the headline and the body text. We also present a data augmentation method that dramatically reduces the text input size a model handles by independently investigating each paragraph of news stories, which further boosts the performance. Our experiments and qualitative evaluations demonstrate that the proposed methods outperform existing approaches and efficiently detect news stories with misleading headlines in the real world.
\end{abstract}

\section{Introduction}

Misleading or false information in journalism has posed a critical social problem~\cite{kwon2013prominent}. Much of the information shared online lacks verification and thus can put our society to unseen threats. News headlines are known to play an important role in making first impressions to readers, and thereby deciding the viral potential of news stories within social networks~\cite{reis2015breaking}. In digital environments under information overload, people are less likely to read or click on the whole contents but just read news headlines~\cite{gabielkov2016social}. Likewise, much of news sharing is headline-based; people circulate news headlines without necessarily having read the full news story. On the other hand, an initial impression gained from the headline is persistent such that its stance remains even after reading the whole news content~\cite{ecker2014effects}. Therefore, if a news headline does not correctly represent the news story  --- or \textit{is incongruent} --- it could mislead readers into advocating overrated or false information, which then becomes hard to revoke.

Identifying incongruent headlines in advance will better assist readers to choose which news stories to consume, and thus decrease the chance of encountering unwanted information. Most previous research tackling this problem has tried to detect incongruity in news headlines either by analyzing linguistic features of news headlines~\cite{blom2015click,chen2015misleading} or by analyzing the textual similarities between news headlines and body text~\cite{ferreira2016emergent,wang2017bilateral}. However, lack of large-scale public dataset makes it difficult to develop sophisticated deep learning models that are better suited for such challenging detection tasks, which usually require million-scale dataset across various domains~\cite{lowe2015ubuntu,go2009twitter}.

This paper, in an attempt to tackle the incongruent news headline problem, presents a million-scale dataset that is built on real news articles published over the course of two years. We propose deep learning approaches that learn the complex textual relationship between the news headline and the full news content, which turned out to be critical for classifying incongruent headlines. Our models were tested on both synthetic data as well as real news stories in the wild. Our contributions are summarized as follows:

\begin{enumerate}
\item 
We release a million-scale dataset for the incongruent headline problem, which covers almost all of the news articles published in a nation over two years. The corpus is composed of pairs of news headlines and body text along with the annotated incongruity label.
\item 
We propose deep hierarchical models that encode the full news article from a word-level to a paragraph-level.
Experiments show our models outperform baseline approaches.
We also present a data augmentation method that splits news paragraphs and annotates each of them separately. This method not only reduces the data size that a model handles but also increases the number of training instances, which further boosts the performance.
\item 
We extensively evaluate our models with real data. Manual verification successfully demonstrates the efficacy of our dataset in training of incongruent headlines. In addition, a crowdsourced experiment suggests the perceived level of incongruence could differ for certain news topics (e.g., politics) by individual beliefs and media outlets.

\end{enumerate}

\begin{table*}[t]
\small
\centering


\begin{tabular}
{C{0.45\columnwidth}C{0.13\columnwidth}C{0.13\columnwidth}C{0.13\columnwidth}C{0.13\columnwidth}C{0.13\columnwidth}C{0.13\columnwidth}C{0.13\columnwidth}C{0.13\columnwidth}C{0.13\columnwidth}}
\hline
  \multirow{2}{*}{ \textbf{Dataset}} & \multicolumn{3}{c}{ \textbf{\# Samples} } & \multicolumn{3}{c}{ \textbf{Headline (Avg.)} } & 
  \multicolumn{3}{c}{ \textbf{Body Text (Avg.)} } \\ \cline{2-10} 
& Train     & Dev.      & Test     & \begin{tabular}[c]{@{}c@{}} \# tokens\end{tabular}  & \begin{tabular}[c]{@{}c@{}} \# chunk\end{tabular}   & \begin{tabular}[c]{@{}c@{}} \# tokens\\ / chunk\end{tabular} & \begin{tabular}[c]{@{}c@{}} \# tokens\end{tabular}    & \begin{tabular}[c]{@{}c@{}} \# chunk\end{tabular} & \begin{tabular}[c]{@{}c@{}} \# tokens\\ / chunk\end{tabular} \\ 
\hline
\textsf{whole} & 1.70M  & 100,000  & 100,000  & 13.71  & 1  & 13.71  & 518.97  & 8.37  & 62.00 \Tstrut  \\ 
\textsf{paragraph}  & 14.20M  & 834,064  & 100,000  & 13.71  & 1  & 13.71  & 62.00  & 2.03  & 30.05 \Tstrut  \\ \hline

\end{tabular}
\caption{
Properties of the dataset.
The chunk in the body text implies paragraphs and sentences for the \textsf{whole} and the \textsf{paragraph} dataset, respectively.
}
\label{table_data_stat}
\end{table*}

\section{Related Work}

There has been a growing interest in detecting false information online. Various fake news datasets have been released for developing AI approaches that make a social impact~\cite{kwon2017rumor}.
One line of studies has focused on detecting \textit{clickbait headlines}, a type of web content that attracts an audience and encourages them to click on a link to a particular web page~\cite{chen2015misleading}. One study~\cite{chakraborty2016stop} released a manually labeled dataset and developed an SVM model to predict clickbait based on linguistic patterns of news headlines. Using this dataset, researchers suggested a neural network approach that measures textual similarities between the headline and the first paragraph~\cite{rony2017diving}. A national-level clickbait challenge was held, where the goal was to identify social media posts that entice its readers into clicking a link~\cite{clickbait_challenge}.

The Fake News Challenge 2017 was held to develop methods that aim to estimate the stance of a news article~\cite{fnc1}. This challenge tackles the {stance detection} problem, whose aim is to estimate the polarity of body text against its headline. This dataset provides 50,000 pairs of headline and body text that were generated from 1,683 original news articles. Each data entry is annotated with one of the four stances: agrees, disagrees, discusses, and unrelated. Several deep learning models have been utilized for the task~\cite{riedel2017simple,chopra2017towards}. Among them, the winning model was a stacked ensemble approach that combines predictions from XGBoost~\cite{chen2016xgboost} based on hand-designed features and a deep convolution dual encoder that independently learns word representations from headline and body text through convolutional neural networks. 


This current research tackles the headline incongruence problem, which is a major kind of misinformation due to the discrepancy in news headline and body text. A previous study proposed co-training approaches for a similar problem of detecting ambiguous headlines from the pair of title and body text~\cite{wei2017learning}. However, the researchers utilized a small set of news articles that are manually labeled and not shared for the public. Until today we notice there has been no million-scale data openly available. 

\begin{figure}[t]
\centering
\includegraphics[width=0.90\columnwidth]{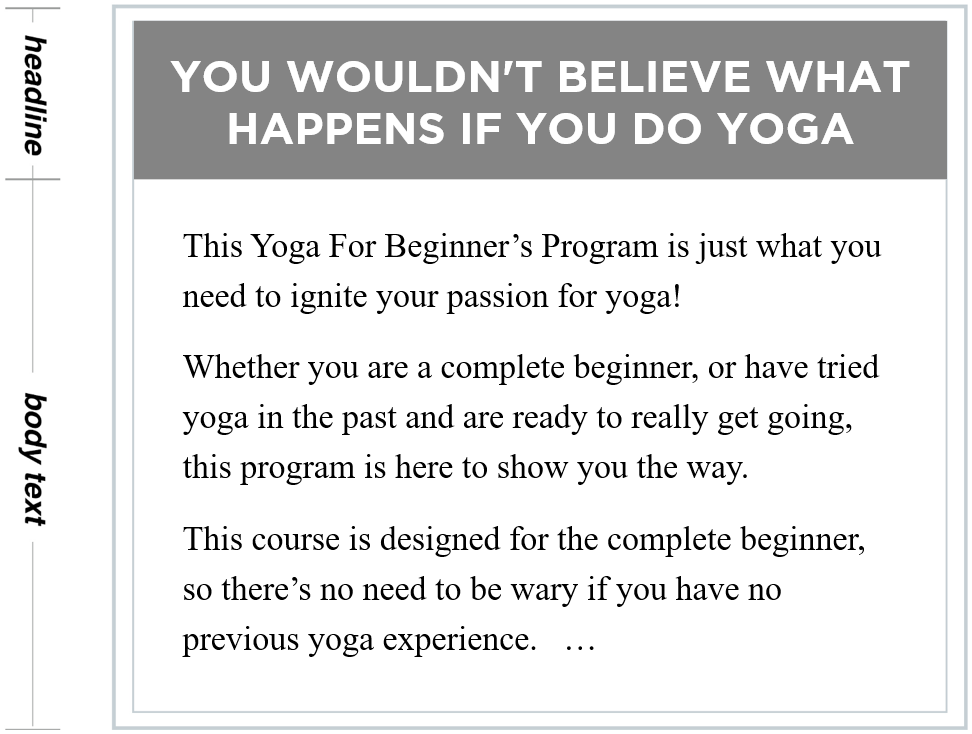}







\caption{
A news article example of the incongruent headline.
}
\label{fig_example_design}
\end{figure}

\section{Problem and Dataset}

The specific problem we tackle is \textbf{the Headline Incongruence Problem}~\cite{chesney2017incongruent}, where a headline of news article holds unrelated or distinct claims with the stories across its body text. Such incongruity in news stories is a major characteristic as clickbait. Figure~\ref{fig_example_design} demonstrates a representative example of such misinformation. The catchy news headline promises to tell certain benefits of yoga, yet the body text mainly is an advertisement for a new yoga program. 
Such incongruent headlines not only make a wrong impression on readers~\cite{ecker2014effects}, but also become worse when it is shared on social media where most users just share without reading its actual contents~\cite{gabielkov2016social}. Therefore, it is crucial to develop automated approaches that detect incongruent headlines in news articles.

To create a new dataset, we crawled a nearly complete set of news articles published in South Korea from January of 2016 to October of 2017. From over 4 million news articles, we performed a series of cleansing steps such as removing non-critical information (e.g., reporter name, non-textual information such as photos and videos). 
Next, we transformed word tokens to integers, which will be released with vocab to help researchers utilize the dataset without the language barrier.

Following the above process, we generated the English-version of dataset based on a large corpus of 0.12m news articles~\cite{horne2018sampling}. In this paper, we do not report any results from the English version of the dataset for brevity, yet we release the two datasets together on this github page~\footnote{http://github.com/david-yoon/detecting-incongruity/} for the research community.

\subsubsection{\textbf{Training Set Creation:} }
\label{create_dataset_whole}
It is almost impossible to manually investigate million-scale news articles for any task. Here, we automatically generated labels on crawled news articles. Rather than crafting new headlines, we implanted unrelated or topically-inconsistent content into body text of original news articles. This process can make a pair of the \{headline\} and \{body text\} where the headline tells distinct stories with its article content. 
Hence, the automation process for creating incongruent-labeled data involves the following steps: (1) sampling a target article from the corpora (which may be on similar topics), (2) sampling part-of-content from another article of the corpora, and (3) inserting this part-of-content to the target article.

We created congruent-labeled data by choosing them from the appropriate corpora. No single headline in this set overlaps with the incongruent-labeled data. Nonetheless, this process may incur false-negative instances when a real article having incongruent headline is chosen inappropriately as a target. We took additional steps to reduce any Type II error via rule-based preprocessing such as inspecting advertising phrases with an n-gram dictionary. We also hired human annotators to manually read 1,000 randomly sampled articles from the created dataset and check whether their headlines are incongruent with the article content. Above manual inspections demonstrated that our method successfully generates news articles where its headline is incongruent with its whole article content. 
We refer to this dataset as \textsf{whole} for comparison with another dataset described below.

\subsubsection{\textbf{Paragraph Set Creation:} }
\label{create_dataset_para}
The task of detecting incongruent headlines can be converted into a set of sub-problems that inspect the textual relationship between a headline and each paragraph respectively, rather than examining the relationship between the headline and whole article content at once. Thus we created the \textsf{paragraph} dataset that transforms a pair of \{headline\} and \{body text\} into multiple sub-pairs of \{headline\} and \{paragraph\}. This conversion process not only reduces the length of text that a model should process but also increase the total number of training instances.
In generating incongruent-labeled data, we sampled \{headline\} and \{paragraph\} from different articles and then matched them as pairs.


In this way, we created {incongruent}- and {congruent}-labeled datasets and maintained train, development, and test datasets that do not overlap each other. All datasets and their detailed description including the original news articles and the indexed version will be made publicly available for the research community.

\section{Methodology}
Our objective is to determine whether a news article contains an incongruent headline, given a pair of \{headline\} and \{body text\}. We call the output probability being incongruent headline \textbf{incongruence score} in this paper.

\subsection{Baseline approaches}

We introduce four baseline approaches that have been applied to the headline incongruence problem.
%
Feature-based ensemble algorithms have been widely utilized for their simplicity and effectiveness. Among various methods, the XGBoost algorithm has shown superior performance across various prediction tasks~\cite{chen2016xgboost}. For example, in a recent challenge on determining the stance of news articles at FNC-1, the winning team applied this algorithm based on multiple features to measure similarities between the \{headline\} and \{body text\}~\cite{fnc_winner}. As a baseline, we implemented the \textbf{XGBoost (XGB)} classifier by utilizing the set of features described in the winning model, such as the cosine similarities between the \{headline\} and \{body text\}. In addition to this model, we also trained \textbf{Support Vector Machine (SVM)} classifiers based on the same set of features.

\subsubsection{\textbf{Recurrent Dual Encoder (RDE)}}
\label{RDE}

A recurrent dual encoder that is consisted of dual RNNs has been utilized to calculate a similarity between two text inputs~\cite{lowe2015ubuntu}. 
We apply this model to the headline incongruence problem via dual RNNs that encode the \{headline\} and \{body text\}, respectively.
When RNN encodes word sequences, each word is passed through a word-embedding layer that converts a word index to a corresponding 300-dimensional vector.
After the encoding step, the probability of being incongruent headline is calculated by using the final hidden state of each \{headline\} and \{body text\} RNNs. 
The incongruence score in the training objective is as follows:

\begin{equation}
\begin{aligned}
& p(\text{label}) = \sigma ( ({h_{t_h}^H})^{\intercal}M~h_{t_b}^B + b ), \\
& \mathcal{L} = -\log \prod_{n=1}^{N} p(\text{label}_n | h_{n,t_h}^H, h_{n,t_b}^B),
\end{aligned}
\label{eq_de_loss}
\end{equation}
where $h_{t_h}^H$ and $h_{t_b}^B$ are last hidden state of each \{headline\} and \{body text\} RNN with the dimensionality $h \in \mathbb{R}^d$. The $M \in \mathbb{R}^{d \times d}$ and bias $b$ are learned model parameters. $N$ is the total number of samples used in training and $\sigma$ is the sigmoid function.

\subsubsection{\textbf{Convolution Dual Encoder (CDE)}}
\label{CDE}
Following the CNN architecture for text understanding \cite{kim2014convolutional}, we apply Convolutional Dual Encoder to the headline incongruence problem. Taking the word sequence of \{headline\} and \{body text\} as input to the convolutional layer, we obtained a vector representation $\boldsymbol{v}= \{ v_{i} | i=1, \cdots, k\}$ for each part of the article through the max-over-time pooling after computing convolution with $k$ filters as follows:
\begin{equation}
\begin{aligned}
& v_i = g(f_i(W)),
\end{aligned}
\end{equation}
where $g$ is max-over-time pooling function, $f_i$ is the CNN function with \textit{i}-th convolutional filter, and $W \in \mathbb{R}^{t \times d}$ is a matrix of the word sequence.
We use dual CNNs to encode the \{headline\} and the \{body text\} into vector representations. After encoding each part of the news article, the probability that a given article has the incongruent headline is calculated in a similar way to the equation (\ref{eq_de_loss}).

\subsection{Proposed methods}
While existing approaches perform reasonably for short text data, dealing with a long sequence of words in news articles will result in degraded performance~\cite{pascanu2013difficulty,bengio1994learning}. 
For example, the recurrent neural network utilized in RDE is poor in remembering information from the distant past.
While CDE learns local dependencies between words, its typical length of convolutional filter keeps the model from capturing any relationship between the words in distinct positions. The inability to handle long sequences is a critical drawback of applying the standard deep approaches to the headline incongruence problem because a news article can be very long. The average word count in our news corpus is 518.97. 

Therefore, we fill this gap by proposing neural architectures that efficiently learn hierarchical structures of long text sequences. We also present a data augmentation method that efficiently reduces the length of the target content while increasing the size of the training set.

\subsubsection{\textbf{Attentive Hierarchical Dual Encoder (AHDE)}}
\label{AHDE}
Inspired by a previous approach that models textual similarity among question-answer pairs using a hierarchical architecture~\cite{yoon2018learning}, this model splits text into a list of paragraphs and encodes the entire text input from the word-level to the paragraph-level via employing a two-level hierarchy of the RNN architecture.
Attention mechanism is added in paragraph-level RNN so that the model can learn the importance of each paragraph in \{body text\} according to \{headline\} of the article. Additionally, we adopt bi-directional RNNs in paragraph-level RNN to exploit information both from the past and the future.

\begin{figure}[t]
\small
\centering
\includegraphics[width=0.85\columnwidth]{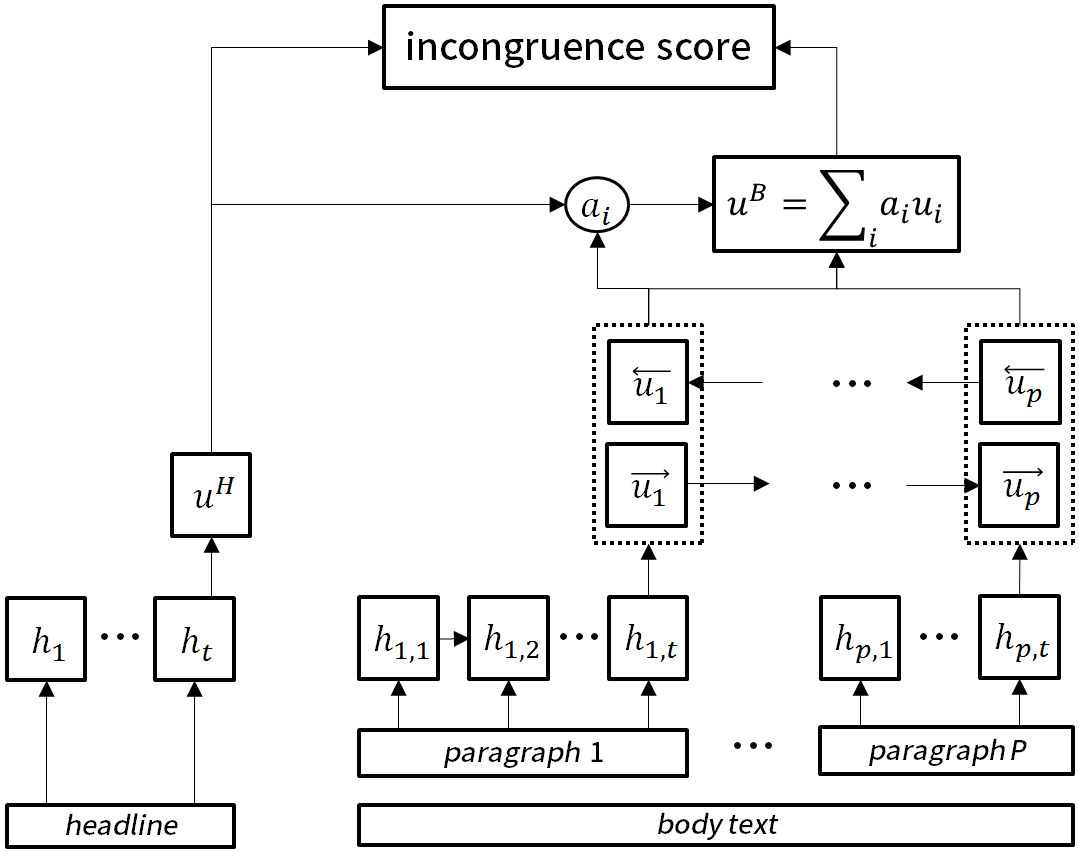}
\caption{
A diagram of the AHDE model.
Entire text input is encoded from the word-level
to the paragraph-level via employing a two-level hierarchy.
The model can learn the importance of each paragraph in body
text according to the headline of the article from an attention mechanism.
}
\label{fig_ahde}
\end{figure}

Figure~\ref{fig_ahde} depicts a diagram of the model.
For each paragraph, the word-level RNN encodes the word sequences $\boldsymbol{w}_p={\{w_{p,1:t}\}}$ to $\boldsymbol{h}_p={\{h_{p,1:t}\}}$. 
Next, the hidden states of the word-level RNN are fed into the next-level RNN that models a sequence of paragraphs while preserving the order. The hierarchical architecture can learn textual patterns of news articles with fewer sequential steps for RNNs compared to the steps required for RDE. While RDE requires an average of 518.97 steps to learn news articles in our dataset, AHDE only accounts for 62.0 and 8.37 steps for each level of RNN, on average.   
The hidden states of hierarchical RNNs are as follows:
\begin{equation}
\begin{aligned}
 &h_{p,t} = f_{\theta}(h_{p,t-1}, w_{p,t}), \\
 &u_p = g_{\theta}(u_{p-1}, h_p),
\end{aligned}
\label{eq_hrde}
\end{equation}
where $u_p$ is the paragraph-level RNN's hidden state at the \textit{p}-th paragraph sequence, and $h_p$ is the word-level RNN's last hidden state of each paragraph $h_p \in\{h_{1:p,t}\}$.
Then, each $u_p$ of \{body text\} is aggregated according to its correspondence with the \{headline\} as follows:
\begin{equation}
\begin{aligned}
 &s_p = \text{v}^\intercal tanh(\text{$W_u^B$}u_{p}^B + \text{$W_u^H$}u^H), \\
 &a_i = \text{exp}(s_i) / {\scriptstyle\sum_p}\text{exp}(s_p), \\
 &{u^B} = {\scriptstyle\sum_i} a_i u_{i}^B,
\end{aligned}
\label{eq_ahde}
\end{equation}
where $u_{p}^B$ indicates the \textit{p}-th hidden state of the paragraph-level RNN that learns the representation of \{body text\}. The $u^H$ indicates the last hidden state of the paragraph-level RNN with the \{headline\}.
We use the same training objective as the RDE model, and the incongruence score is calculated as follows:
\begin{equation}
\begin{aligned}
& p(\mbox{label}) = \sigma ( ({u^H})^\intercal M~u^B + b )
\end{aligned}
\label{eq_ahde_loss}
\end{equation}

\vspace*{1mm}
\subsubsection{\textbf{Hierarchical Recurrent Encoder (HRE)}}
\label{HRE}
The AHDE model uses two hierarchical RNNs for encoding text from word-level to paragraph-level. The model requires higher computation resources in training and inferencing compared to those of non-hierarchical alternatives such as RDE and CDE. Therefore, we investigate an intermediate approach that models hierarchical structures of news articles with a simpler neural architecture.
The text in the news \{body text\} is split into paragraphs, each of which is embedded by averaging the word-embedding vector from its containing words.
In other words, 
HRDE calculates $h_{p}$ in equation (\ref{eq_hrde}) by averaging the word embedding among the words in paragraph $p$, $h_p = \sum_i embedding(w_i),~w_i\subset{\mbox{\textit{p}-th~paragraph}}$. Then, paragraph-level RNN is applied with the paragraph-encoded sequence input, $h_{p}$, for retrieving the final encoding vector of the whole \{body text\}.

\begin{figure}[t]
\small
\centering
\includegraphics[width=0.9\columnwidth]{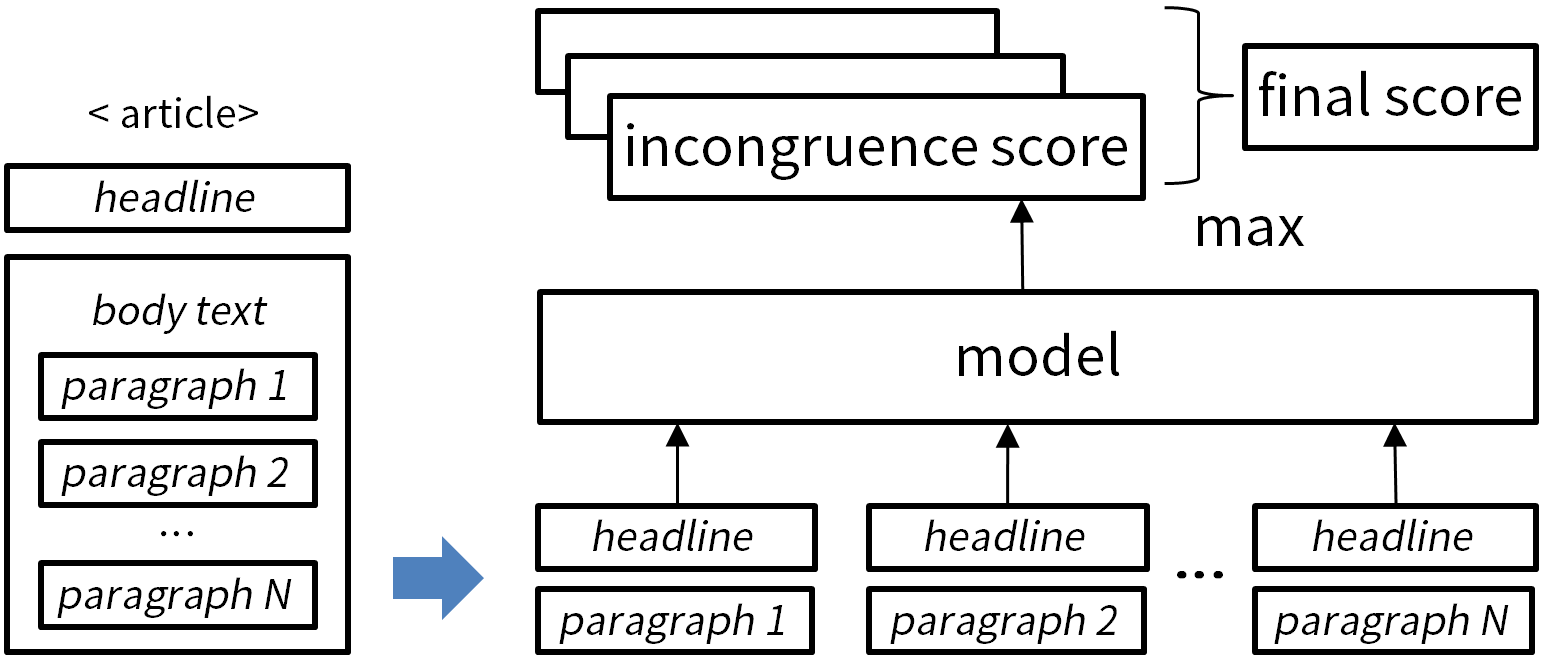}
\caption{
Diagram of the independent paragraph method. A given news article is split by its paragraphs, each of which is compared to the headline to calculate incongruence score. The maximum value is taken as the final incongruence score.}
\label{fig_ip}
\end{figure}
\begin{figure*}[t]
\small
\centering
\subfigure[without IP method]{\label{fig_analysis_length_whole}\includegraphics[width=0.99\columnwidth]{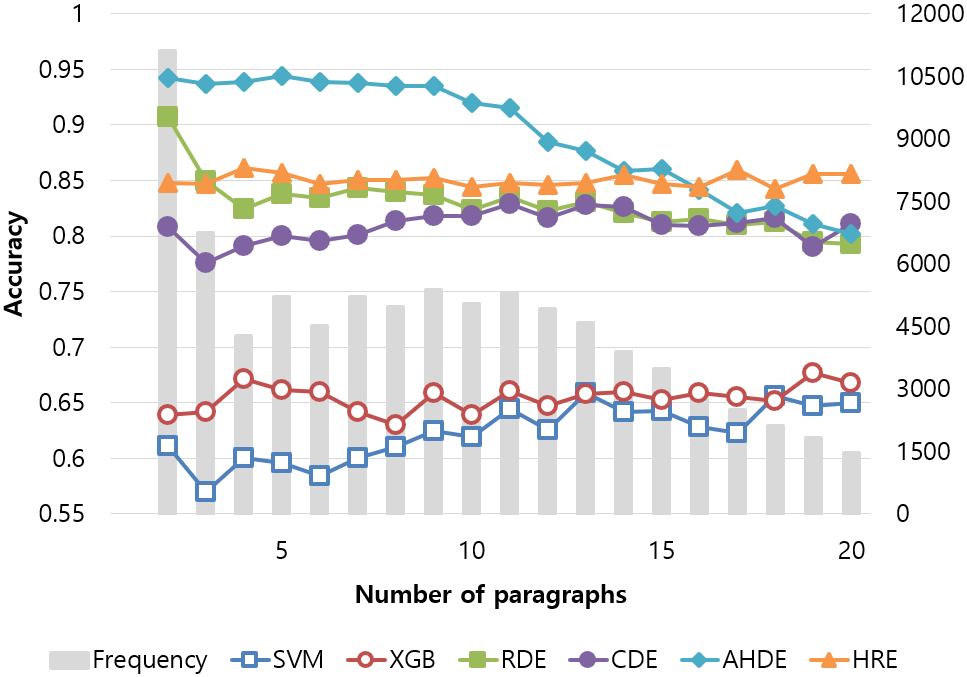}}
\qquad
\subfigure[with IP method]{\label{fig_analysis_length_para}\includegraphics[width=0.99\columnwidth]{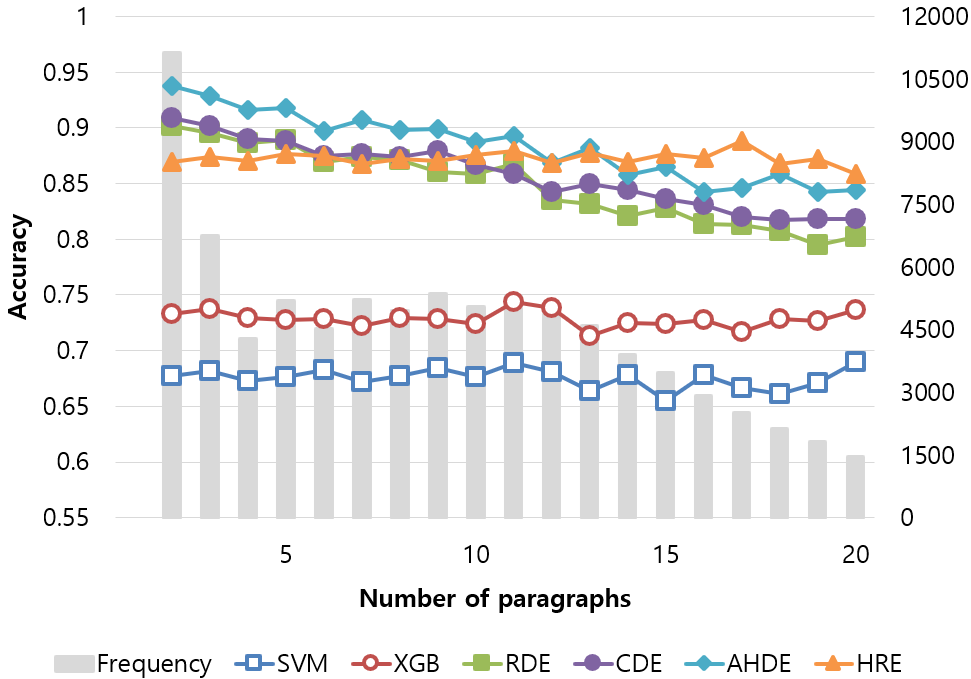}}
\caption{
Prediction performances across a different number of paragraphs in body text of the test dataset (a) without Independent Paragraph (IP) method and (b) with IP method. Line plots demonstrate prediction accuracies with increases in the number of paragraphs, and gray bars present the frequency of the test instances having a same number of paragraphs.
}
\label{fig_analysis_length}
\end{figure*}

\subsubsection{\textbf{Independent Paragraph (IP) Method}}
\label{ip_method}
In addition to the neural architecture, we propose a data augmentation method that splits paragraphs in the \{body text\} and learns the relationship between each paragraph and headline independently.

Figure~\ref{fig_ip} depicts the diagram of the IP method, which computes \textbf{incongruence score} for each paragraph from its relationship with the news headline. 
The final incongruence score for the pair of \{headline\} and \{body text\} is determined as the maximum score of incongruence scores as follows:
\begin{equation}
\begin{aligned}
	& p(\mbox{label}) = max(s_{1:p}), \\
\end{aligned}
\label{eq_ip}
\end{equation}
where $s_p$ is the incongruence score calculated from the \textit{p}-th paragraph of the \{body text\} and \{headline\}. 
The selection of the maximum score can better identify news articles which contain a paragraph that is highly unrelated to the news headline.

For training models with IP method, we use the \textsf{paragraph} dataset and incongruence scores are calculated in the following ways:
\begin{itemize}
\item \textbf{XGB/SVM with IP:} Extracting features from \{headline\} and each paragraph of \{body text\}, XGB/SVM measures the incongruence score for each paragraph. 

\item \textbf{RDE/CDE with IP:} Both models encode word sequences in each paragraph of \{body text\} and compare them with the encoded \{headline\}.

\item \textbf{AHDE with IP:} To encode each paragraph in \{body text\}, the first-level RNN encodes word sequences for each sentence and the second-level RNN takes a sequence of sentences as input, which is retrieved from the first-level RNN.

\item \textbf{HRE with IP:} To obtain the incongruence score for each paragraph, HRE first calculates the mean of word vectors for each sentence. Then, RNN encodes a sequence of sentences by taking the averaged word vectors as input.
\end{itemize}

\begin{table}[t]
\small
\centering
\vspace{-2mm}
\begin{tabular}{C{0.2\columnwidth}C{0.12\columnwidth}C{0.14\columnwidth}C{0.12\columnwidth}C{0.14\columnwidth}}
\hline
\multirow{2}{*}{Model} & \multicolumn{2}{c}{Without IP} & \multicolumn{2}{c}{With IP} \\ \cline{2-3} \cline{4-5} 

 	& Acc.\Tstrut   &AUROC\Tstrut &Acc.\Tstrut &AUROC\Tstrut
\\ \hline 
SVM\Tstrut		&0.640\Tstrut	&0.703\Tstrut	&0.677\Tstrut	&0.809\Tstrut	\\
XGB		&0.677	&0.766	&0.729	&0.846	\\
CDE		&0.812	&0.900	&0.870	&\textbf{0.959}	\\
RDE		&0.845	&\textbf{0.939}	&0.863	&0.955	\\
\hline
AHDE\Tstrut	&\textbf{0.904}\Tstrut	&\textbf{0.959}\Tstrut	&\textbf{0.895}\Tstrut	&\textbf{0.977}\Tstrut	\\
HRE		&\textbf{0.850}	&0.927	&\textbf{0.873}	&0.952	\\   \hline

\end{tabular}

\caption{
Model performance (top-2 scores marked as bold).
}
\label{table_accuracy_auroc}
\end{table}

\section{Experiments}

We conduct a series of experiments to compare baseline methods with the newly proposed models. All codes developed for this research will be made available via a public web repository along with the dataset. 
We provide further implementation details in the Appendix, which are necessary to reproduce the results in this paper.

\subsection{Performance Comparison}

Table~\ref{table_accuracy_auroc} presents performances of all approaches. Performances using \textsf{whole} dataset is also compared with those with the IP method on \textsf{paragraph} dataset. We report accuracy and the AUROC (Area Under Receiver Operating Characteristic) value, which is a balanced metric with regard to the label distribution. 

We first find that the four deep learning models 
outperform feature-based machine learning models.
Among the deep learning models, the newly proposed AHDE achieved the best performance with regard to accuracy and AUROC (0.904 and 0.977, respectively). Second, prediction performance increased significantly when the IP method was applied. RDE and CDE got the advantage of the IP method most, such that they even showed the performances comparable to the hierarchical models.
Even though those simple models do not have an appropriate structure to handle lengthy news data (i.e., news posts and news paragraphs on average contain 518.97 and 62.00 words respectively in Table~\ref{table_data_stat}), the IP method did help them examine the relationship between the headline and each paragraph more efficiently. 


\subsection{Performance over Long Text Input}

One major limitation of standard deep learning approaches is difficulty in processing very long text. To verify the ability of our models in handling lengthy news articles, we measured the model performance under increasing input size. Figure~\ref{fig_analysis_length} shows the result. 

First, as observed in Figure~\ref{fig_analysis_length_whole}, newly proposed models (AHDE and HDE) showed consistently higher performance over baseline approaches (e.g., XGB, RDE). When the IP data augmentation method was applied, all six models benefit and show an increase in performance as shown in Figure~\ref{fig_analysis_length_para}. Second, the figure suggests the robustness of our proposed models in handling long sequential input by their own hierarchical structures. Whether using the IP method or not, the newly proposed HRE and AHDE model consistently showed competitive performance irrespective of the paragraph size in \{body text\}. 
While AHDE performs better than HRE when the number of paragraphs is a few, the performance gap became narrower as paragraph size increases and HRE achieves the best score for extremely long input (i.e., a news article containing 19-20 paragraphs). This trend could be explained by the fact that HRE has a fewer number of trainable parameters compared to AHDE.

Due to space limitation, we present detailed results that describe model performance over varying content types in Appendix.

\begin{figure}[t]
\small
\centering
\includegraphics[width=0.99\columnwidth]{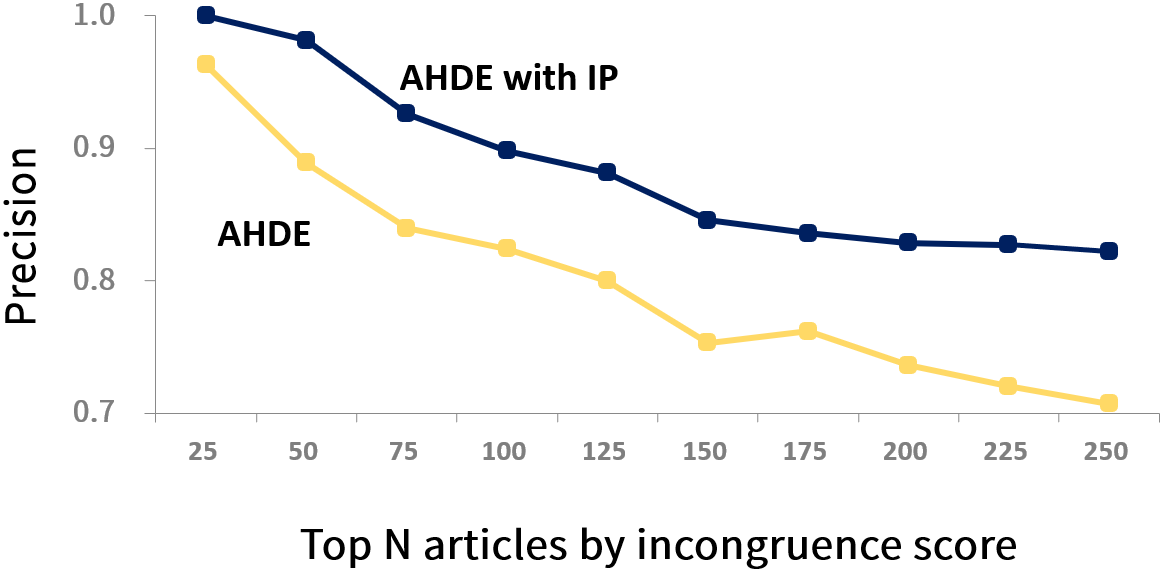}
\caption{
Precision values for detecting news articles with incongruent headlines in the newly gathered dataset.
The x-axis shows the top-N articles by incongruence scores, and the y-axis presents its corresponding precision.
}
\label{fig_real}
\end{figure}

\subsection{Real World Evaluation}

To see the efficacy of our dataset and proposed models for detecting incongruent headlines in the wild, we evaluated our pre-trained models on more recent news articles. We newly gathered 232,261 news articles that were published from January to April of 2018. Testing our model with the newly gathered dataset can show the generalizability of our approach in the real world.

At first, we manually inspected random samples of news articles to see whether they have incongruent headlines. Yet, we could not retrieve enough number of articles having incongruent headlines for evaluation. This is possibly due to the sparse number of incongruent headlines in the real world. Therefore, instead of looking into the randomly sampled dataset and labeling them for evaluation, we decided to manually validate top $N$ articles by incongruence scores that are given by model prediction. Since models give incongruence scores (i.e., output probability) based on its confidence for classification, we believe such evaluation successfully estimates precision scores of prediction models. This type of evaluation is widely used in the tasks where it is impossible to count the true cases in a dataset such as question answering system~\cite{ferrucci2012introduction}. 

Figure~\ref{fig_real} shows the precision scores for AHDE models that are trained with- and without the IP method, respectively. The x-axis presents the top-N articles by incongruence scores that are retrieved by the models out of the newly gathered articles over 4 months. The y-axis demonstrates the precision value corresponding to the top N articles. 

Here we make three observations. First, the AHDE model with the IP augmentation consistently shows higher precision than the AHDE model without the IP method. This finding supports the superior performance of the IP method across different evaluations. Second, the AHDE model with IP achieved the precision of 1.0 for the top 25 articles. Even though the model was trained by a separate dataset, it successfully filtered out real cases where its headline conveys different stories with associated body text.
Third, when we evaluate the top 250 articles, the precision of the AHDE model with IP reduced to 0.82. Nevertheless, this precision value is high enough to be utilized for detection in real news platforms.

The above observations suggest that our approach and dataset could be applied for the headline incongruence problem in the wild. Based on application scenarios, one could control the trade-off between precision and recall by changing the model thresholds.

\section{Discussion}

\subsection{Varying perceptions on headline incongruence}


So far, we have treated incongruence score as an inherent value that is fixed for each news article. We conducted additional surveys using the Amazon Mechanical Turk (MTurk) service to understand whether the general public would also consider the news articles predicted by our models to contain incongruent headlines. We also tested whether people's perception on incongruence score varies by individuals' partisanship, hypothesizing from a previous finding that people's perception on the veracity of news varies by political stance~\cite{allcott2017social}.

We first manually gathered news articles from two media outlets. In order to retrieve as many incongruent news headlines as possible, we selected two media outlets that are considered not trustworthy by common journalistic standards (referring to {mediabiasfactcheck.com}): one was chosen from conservative media (\textsf{Media A}) and another from liberal media (we call \textsf{Media B}). We do not reveal these media names, as the particular choice of a media outlet is less of a concern to our study.
Given the definition of incongruent headline and article selected by the model from each media, we asked 100 Amazon Mechanical Turk workers to answer the following question \textit{``Do you think the headline of the above article is incongruent with its body text?''} 

\begin{figure}
  \centering
  \small
  \hspace*{-5mm}
   \includegraphics[width=1.1\linewidth]{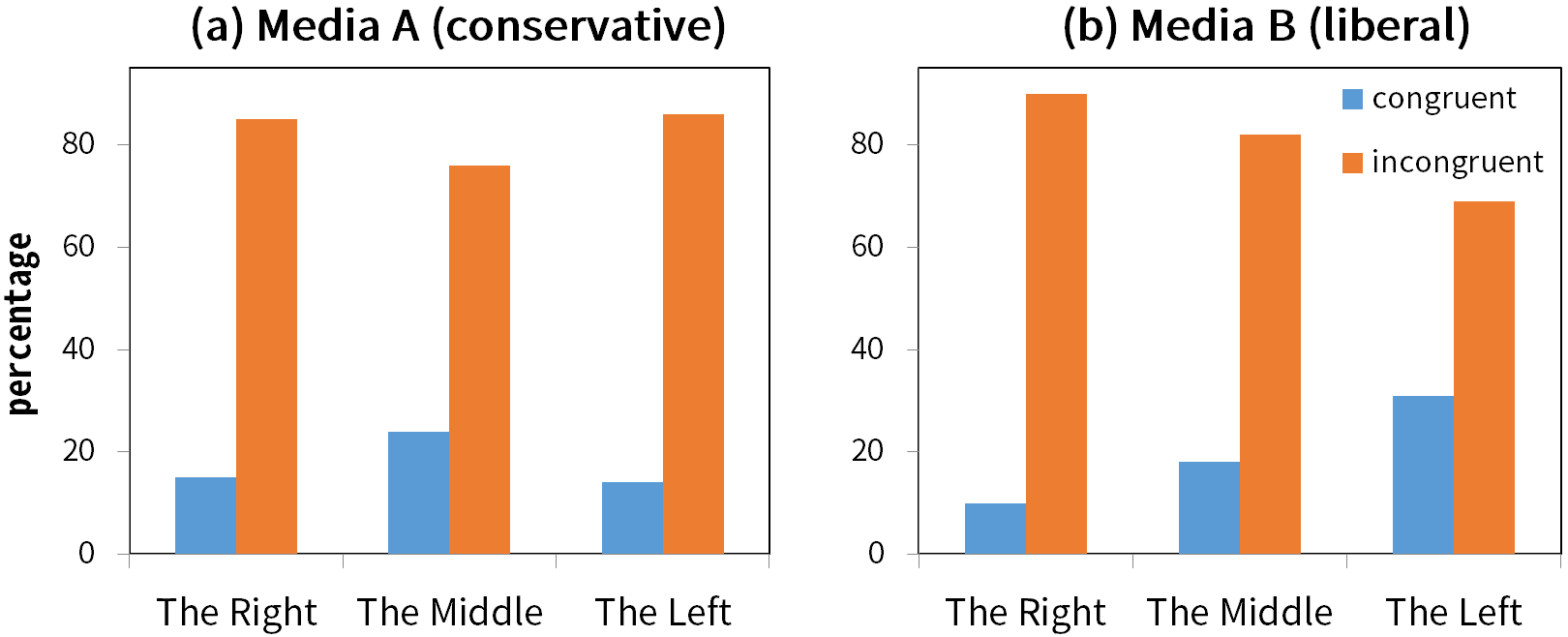}
   \caption{MTurk results indicating political stances of survey participants (the x-axis) and their responses to articles of high incongruence score (the y-axis). }
   \label{fig:result_amt}
\end{figure}

Figure~\ref{fig:result_amt} confirms that MTurk workers tend to find articles of high incongruence scores to contain misleading headlines. One interesting trend to note is the change in perceived incongruence score by individual belief. While non-liberal participants considered news samples in \textsf{Media B} to have a similar level of incongruence to \textsf{Media A} samples, liberal participants found \textsf{Media B} to be less incongruent. \kwm{This finding suggests that while our approach is applicable in general the perceived incongruence level may be judged differently for certain news topics (like politics).} 
\kwm{This implies that news service providers should be cautious when employing human coders and crowdsourcing workforce to get a fair label on misinformation and fake news.}

\subsection{Hierarchical Encoders for Stance Detection}

To further the generalizability of deep approaches proposed in this paper, we conducted an additional experiment on the FNC-1 dataset~\cite{fnc1}, aimed for \textit{stance detection}. This problem is similar to the headline incongruence problem in that one needs to compare the textual relationship between news headline and its whole content, but different in that its target label consists of four different cases (i.e., unrelated, agree, disagree, and discuss). To have a similar setting with our task, we transformed these four labels into binary, which is ``unrelated'' and ``others'' and trained models. 

We compared our hierarchical deep learning approaches (i.e., AHDE, HRE) with feature-based approaches and standard deep learning models. We also included ensemble models that combine the predictions of XGB and each deep learning models, because an ensemble of XGB and CDE was the winning model of the FNC-1 challenge~\cite{fnc_winner}. 
Among single models, XGB outperformed the other models with the accuracy of 0.9279. Among deep learning models, the AHDE model was the best with the accuracy of 0.8444. Better performance of XGB over deep approaches might be caused by insufficient variations of training instances in the FNC-1 dataset. We noticed that even though the training set includes around 50k instances, many news articles of the unrelated label were generated from 1,683 original news articles by swapping headlines with one another and thus 29.7 instances had identical body text.

Above reasons might lead the challenge winners to use ensemble models that combine predictions of feature-based approaches and deep neural networks. The XGB+CDE ensemble model achieved the accuracy of 0.9304, outperforming all of the single models. When we combined the prediction of AHDE with XGB, the ensemble model achieved the best accuracy of 0.9433. Incorporating with the results in Table~\ref{table_accuracy_auroc}, this finding suggests that the proposed hierarchical neural networks effectively learn textual relationships between two texts rather than standard approaches. 
We strongly believe the highest accuracy of XGB among single models is due to the aforementioned limitation of the FNC-1 dataset, and hence the ensemble approach may not be necessary if given dataset is large enough to train neural networks. In additional experiments on our dataset proposed in this paper, we found that the AHDE model solely achieved the best accuracy than any combinations of other approaches for the ensemble.


\subsection{Future Directions}

\kwm{A natural extension of this study is to develop and improve prediction models for detecting news articles with incongruent headlines by considering syntactic features. A simple way is to apply NLP pipelines such as part-of-speech tagging or named entity recognition to reduce the complexity of raw text input. It would be also possible to develop tree-shaped deep neural architectures similar to LSTM-tree~\cite{tai2015improved} to understand the textual relationship between headline and body text more sophisticatedly.}

\kwm{Another study could extend this work to measure the incongruence of title and content across other types of online contents. The title plays a crucial role in attracting users into clicking and consume digital contents such as blog articles, online videos, and even scientific papers. Similar to the incongruent headline problem, automatically identifying such incongruent titles of various contents will assist people to be happier. To this end, future researchers could share different types of the dataset and improve AI approaches that measure the incongruity of title and content.}

\section{Conclusion}
In this paper, we study the problem of incongruent headline detection. 
To this end, we release a million-scale data corpus suitable for detecting news articles where its body text disagrees with the headline. We also propose two neural networks that efficiently learn the textual relationship between headline and body text via a hierarchical recurrent architecture. The experiments demonstrate that the models trained on our released corpus show decent performances both on the synthetic dataset and on the real-world dataset. Moreover, we introduce a data augmentation method, called Independent Paragraph, that makes headline-paragraph pairs by splitting the whole body text into separate paragraphs which improves the performance of the models.
We hope that our dataset and approach will help detect news articles with a misleading headline that can bait readers' attention and hence contribute to the emerging threats of misinformation to our society.

\section{Acknowledgments}
Seunghyun Yoon and Kunwoo Park equally contributed to this work. Majority of this work was done while Kunwoo was at KAIST. We sincerely thank Taegyun Kim for preparation of dataset and the reviewers for their in depth feedback that helped improve the paper. This research was supported by the Ministry of Trade, Industry \& Energy (MOTIE, Korea) under Industrial Technology Innovation Program (No. 10073144), the National Research Foundation of Korea (NRF) funded by the Korea government (MSIT) (No. 2016M3C4A7952632), Basic Science Research Program (No. NRF-2017R1E1A1A01076400) and Next-Generation Information Computing Development Program (No. NRF-2017M3C4A7063570) through the National Research Foundation of Korea (NRF) funded by the Ministry of Science, ICT.

\bibliographystyle{aaai}
\bibliography{aaai19_main}

\appendix

\section{Implementation Details}
\label{more_implementation}

\subsection{Recurrent Dual Encoder (RDE)}
Among variants of RNN functions, the Long Short-Term Memory (LSTM) architecture is commonly used for its efficiency in addressing the exploding and vanishing gradient problem that arises from standard recurrent units~\cite{hochreiter1997long}. Instead of that, we used the Gated Recurrent Unit (GRU), because it shows compatible performance to the LSTM with a fewer number of weight parameters~\cite{chung2014empirical}. We used a two single-layer GRU with 300 hidden units. The \{headline\} and \{body text\} were independently encoded using dual GRUs, which share weights. The hidden states were initialized using orthogonal weights \cite{saxe2013exact}, and the embedding layer was randomly initialized from the Gaussian distribution with 300 dimensions. The vocabulary size in the news dataset was 253,067. 
We used the Adam optimizer~\cite{kingma2014adam} including gradient clipping by norm at a threshold of 1.
For the purpose of regularization, we applied Dropout~\cite{srivastava2014dropout} with the ratio of 0.2.

\subsection{Convolution Dual Encoder (CDE)}
We used two single-layer CNNs with a total of 1,000 filters, which involved five types of filters $K \in \mathbb{R}^{\{1,3,5,7,9\} \times d}$, 200 per type. The weight matrices for the filters were initialized using the Xavier method \cite{glorot2010understanding} and weights for the two CNNs were shared.
Adam optimizer~\cite{kingma2014adam} was used with the initial learning rate of 0.001. Other implementation details are similar to the RDE model. \\

\subsection{Attentive Hierarchical Recurrent Dual Encoder (AHDE)}
We used two single-layer GRU with 300 hidden units for the word-level RNN portion to encode the word sequence in each \{headline\} and \{body text\}. We used another two single-layer bidirectional GRU with 100 hidden units for the paragraph-level RNN portion to encode the final hidden state sequences from each word-level RNN of the \{headline\} and \{body text\}. 
The weights of the GRU are shared within the same hierarchical portion, word-level and paragraph-level. For regularization, dropout was applied with the ratio of 0.3 for the word-level RNN portion in AHDE. The other settings were the same with RDE.

\subsection{Hierarchical Recurrent Encoder (HRE)}
The dimension of the word-embedding vector was set to be 300, which was pre-trained from the training corpus with the skip-gram model~\cite{mikolov2013distributed}. Parameter settings for the paragraph-level RNN in the HRE model were same as those of the AHDE model.

\section{More Analysis of Models for Types of Articles}
\label{more_analysis}

In order to cover various cases in which incongruent headlines can occur, we put three different cases for the generation as described below. In all cases, we set the portion of implanted part-of-contents to take up less than 50\% of the article length.

\begin{enumerate}
\item \label{datatype-1}Sample $n$ consecutive paragraphs from an article and insert them into the \{body text\} of the target article. 

\item \label{datatype-2}Sample $n$ non-consecutive paragraphs from one article and insert them randomly into the \{body text\} of the target article. 
\end{enumerate}

\begin{figure}[t]
\small
\centering
\subfigure[without IP method]{\label{fig_analysis_model_4types_whole}\includegraphics[width=0.9\columnwidth]{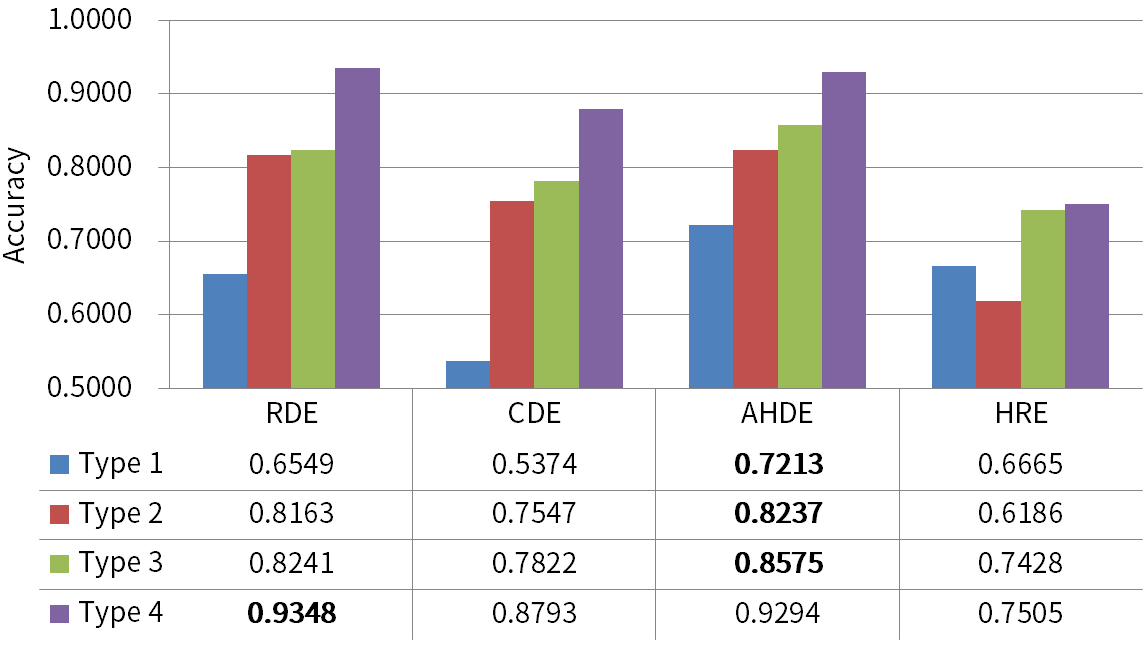}}
\subfigure[with IP method]{\label{fig_analysis_model_4types_para}\includegraphics[width=0.9\columnwidth]{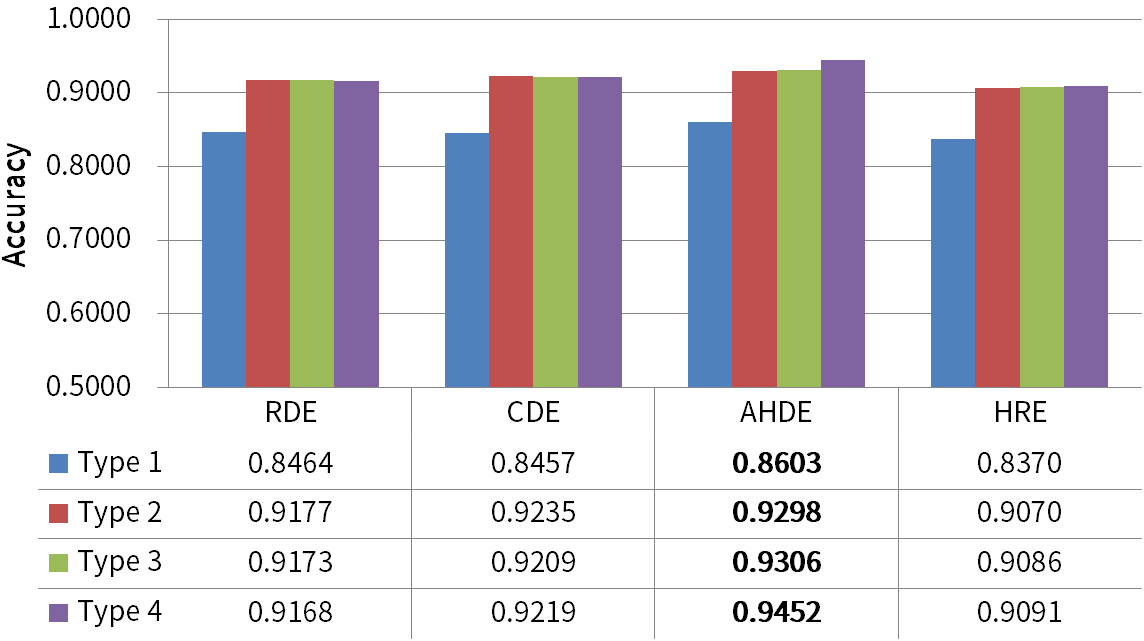}}
\caption{
Analysis of the models. Top performance is marked as bold for each type.
}
\label{fig_analysis_model_4types}
\end{figure}

To better understand how the models work, we used four different types of test datasets generated separately by the rule 1 and 2.
\begin{itemize}
\item  \textbf{Type 1}: Applying rule (1) with only one paragraph ($n$ = 1)
\item  \textbf{Type 2}: Applying rule (1) with two or more consecutive paragraphs ($n >$ 1)
\item  \textbf{Type 3}: Applying rule (2) without random arrangement (maintaining the ordering of the sampled paragraphs) ($n >$ 1)
\item  \textbf{Type 4}: Applying rule (2) ($n >$ 1)
\end{itemize}
The difference between the Type 1 and Type 2 data lies in the number of paragraphs implanted into a target article. 
Type 2 data represents the cases that contain a larger portion of implanted contents.
With comparisons of Type 3 and Type 4 data, we could understand how the models perform when they examine an article that has implanted chunks of text scattered in the \{body text\} whether keeping the original order of its context or not.
%
The evaluation results of each model with the 4 types of test dataset are presented in Figure~\ref{fig_analysis_model_4types}.
\vspace{0.15cm}

\vspace*{1mm}
\noindent\textbf{AHDE/RDE: }
These RNN-based approaches performed better in all types compared to the others, which is attributed to the ability to use sequential information. In our experiments, when implanted contents appeared at the beginning or the end, AHDE and RDE were highly accurate (`without IP method' case, 0.15 - 0.2 higher than average), while the others yielded scores similar to their averages. Furthermore, these models achieved a relatively high-performance gain by randomly shuffling the order of implanted contents (comparing Type 3 with Type 4 data). Additionally, we can see the effect of the hierarchical architecture by comparing Type 1 to Type 2 data. AHDE and RDE achieve similar accuracy for Type 2 data, yet AHDE performed better on Type 1 data (i.e., fewer polluted sentences). This indicated that the model could memorize the article more systematically via the hierarchical architecture.

\vspace*{1mm}
\noindent\textbf{CDE: }
CDE is a CNN-based model that uses a specific window of words as contextual information. Local features are computed using filters of various sizes and the model finds the max value over time. In this way, CDE extracts one of the most effective features to determine whether the article has incongruent headlines. Thus CDE depends upon the number of incongruent words, rather than the patterns of implanted paragraphs, such as the order or the insertion position, to investigate the relationship between headline and body text. In Type 2, CDE produced results comparable to the others. This is because the article has more numbers of incongruent words, so there is a higher chance of the model extracting useful features. However, this is not guaranteed when implanted contents covered a similar topic.

\vspace*{1mm}
\noindent\textbf{HRE: }
HRE performance is degraded when the portion of implanted news contents becomes larger. This is because the number of commonly used words increases with the sentence length. This leads to a dilution of the detection features on the word embedding, and then the model does not work well. As with AHDE/RDE, HRE is an RNN-based model, which means it also utilizes sequential information. This effect is particularly noticeable when implanted paragraphs are scattered throughout the article (as in Type 3 and Type 4 data).

\vspace*{1mm}
\noindent\textbf{With Independent Paragraph method: }
For each type, we have conducted experiments using both with- and without-IP methods.
From Figure~\ref{fig_analysis_model_4types_whole} and \ref{fig_analysis_model_4types_para}, we find that the IP method improves the performance compared to the without-IP method and such a trend is particularly pronounced when the portion of implanted content is relatively small. 
This is because when news content is divided into several paragraphs, the model can analyze each part in more detail. The hierarchical architecture enables the model to produce similar effects even on the whole text. With multiple systematic encoding, it can help the model remember the lengthy or complex things that might be missed when using a simple model.

\end{document}